\documentclass{article}
\usepackage{spconf}
\usepackage[cmex10]{amsmath}
\usepackage{amsthm}
\usepackage{amssymb}
\usepackage{mathrsfs}
\usepackage{graphicx}
\usepackage{float}
\usepackage{array}
\usepackage{epstopdf}
\usepackage{multirow}
\usepackage{environ}
\usepackage{relsize}

\newcommand{\argmin}{\arg\!\min}

\title{M\MakeLowercase{ultispectral} P\MakeLowercase{almprint} R\MakeLowercase{ecognition} U\MakeLowercase{sing} \MakeLowercase{A} H\MakeLowercase{ybrid} f\MakeLowercase{eature}}
\name{}
\name{Sina Akbari Mistani, Shervin Minaee and Emad Fatemizadeh}
\address{Electrical Engineering Department, Sharif University, Tehran, Iran.}
\allowdisplaybreaks[1]	
\begin{document}
%
\maketitle
\begin{abstract}
Personal identification problem has been a major field of research in recent years. Biometrics-based technologies that exploit fingerprints, iris, face, voice and palmprints, have been in the center of attention to solve this problem. Palmprints can be used instead of fingerprints that have been of the earliest of these biometrics technologies. A palm is covered with the same skin as the fingertips but has a larger surface, giving us more information than the fingertips.
The major features of the palm are palm-lines, including principal lines, wrinkles and ridges. Using these lines is one of the most popular approaches towards solving the palmprint recognition problem. Another robust feature is the wavelet energy of palms.
In this paper we used a hybrid feature which combines both of these features. 
At the end, minimum distance classifier is used to match test images with one of the training samples. 
The proposed algorithm has been tested on a well-known multispectral palmprint dataset and achieved an average accuracy of 98.8\%. 
\end{abstract}

\section{Introduction}
Personal identification has always been a crucial issue in critical tasks and personal devices.
Conventional methods to this end include passwords and ID cards. However, besides common problems with memorizing passwords and maintaining an ID card, they are exposed to being disclosed or stolen, threatening the security.
Today the area of personal identification is exploiting computer-aided systems as a safer and more robust method, and biometrics are among the most reliable features that can be used in computer-aided personal recognition.
Inconveniences associated with the older methods caused a rapid peak in the popularity of biometric-based applications.
The most widely-used biometrics include fingerprints \cite{fing}, facial features \cite{face}, iris patterns \cite{iris} and palmprints \cite{palm}.
Palmprints provide a number of privileges over other biometric features, making them an appropriate choice for identification applications. First, they are more economical, as palmprint images can be easily obtained using inexpensive CCD cameras.
Second, they are robust as hand features do not change significantly over time.
Finally, palmprint-based systems work well in extreme weather and illumination conditions, since palmprints contain much richer set of features than many other biometrics.

There are several useful features in a palm image such as line features (principal lines and wrinkles), geometrical features such as size of the palms, the angles between principal lines, textural features, etc.

Although line-based features are very useful in palmprint recognition, systems that work only based on these features face a few common dilemmas.
One such issue is that in some cases principal lines and wrinkles are not enough to discriminate palms since there are many palms with similar line features. Fig.1 shows two palms with similar line pattern.
\begin{figure}[2 h]
\begin{center}
    \includegraphics [scale=0.6] {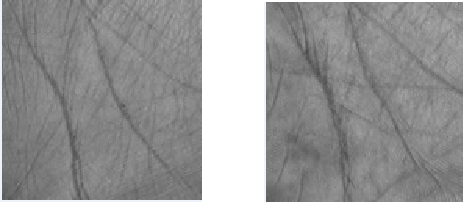}
 \end{center}
 \vspace{-0.2cm}
  \caption{ Two different palms with similar principal lines}
\end{figure}
\\
Another common problem is that in many cases palmprint lines are difficult to extract because of the low quality of database images; two such palmprint images are shown in Fig.2.
\begin{figure}[3 h]
\begin{center}
    \includegraphics [scale=0.6] {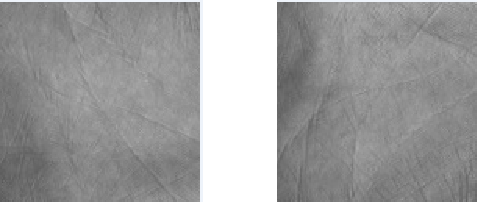}
 \end{center}
 \vspace{-0.2cm}
  \caption{ Palmprints with unclear principal lines}
\end{figure}

One useful way to overcome these problems and to improve the accuracy of palmprint identification systems is to use multispectral imaging \cite{multi_spectral}, which captures an image in a variety of spectral bands (usually four in palmprint recognition). Each spectral band highlights specific features of the palm and helps us derive more information from the palmprint. 
So far, many approaches have been proposed for palmprint recognition. In \cite{kong}, a survey of palmprint recognition systems is provided. The author has divided most of the previous algorithms into the following categories. There are some texture-based approaches (most of which use statistical methods for feature extraction). Many previous approaches are based on features derived from transform domains such as Gabor, Fourier and wavelet transforms. As an example in \cite{wavelet_fuse}, Han used a wavelet-based image fusion method for palmprint recognition.
There are also several line-based approaches developed for palmprint recognition. Palm lines including principal lines and wrinkles have proven very useful in this area. In \cite{wu}, Wu proposed multiple line features and used them for palmprint matching.
Another group of methods use image coding techniques to perform palmprint recognition. In \cite{orient}, Jia used a robust line orientation code for palmprint verification. Some coding methods are also used for palmprint recognition, such as palm code, fusion code, competitive code and ordinal code \cite{code}. In \cite{QPCA}, Xu proposed a quaternion principal component analysis approach for multispectral palmprint recognition which achieved a high accuracy rate. Ekinci \cite{PCA+GWR} proposed a Gabor wavelet representation approach followed by kernel PCA for palmprint recognition.

As discussed before, palm-line extraction could be difficult for some images. The reason of this difficulty lies in the similarity of the principal lines, reducing their discrimination capability from other palms. To overcome these problems, we propose a hybrid feature that uses both the principal line information and also local wavelet energy of palms. 
This hybrid feature combines the following two features:
\vspace{-0.1cm}
\begin{enumerate}
\item	Principal lines and their energy in different locations of the palm.
\item	Wavelet transform of palm image that can help us detect the small differences between different palms.
\end{enumerate}
\vspace{-0.01cm}
These two features are explained in more details in Section 2.1 and Section 2.2 respectively.
In Section 3 we explain how to combine them in a proper way by scaling them in the same range. Section 4  provides the experimental results of this algorithm on PolyU multispectral dataset \cite{dataset} and Section 5 provides some conclusions.

\section{Palmprint features}
In this paper, we developed a recognition algorithm for multispectral palmprint images.
Multispectral methods require different samples of the same object in order to make a better decision. Here we assume that in the image acquisition step, four images of each palmprint are acquired using four CCDs. These images are then preprocessed and the regions of interest (ROI) for each of them are extracted. Provided these ROIs, features are extracted from each palm. Basically we employ two parallel feature extraction paths and the final decision is  is made based upon the results of the two paths.
The two features used in this paper are local palm-line energy and local wavelet energy of palmprint. The basic idea is to divide palm images into a number of non-overlapping blocks and then extract the energies of all blocks as features.

\subsection{Palm Line Extraction}
As mentioned earlier, palmprints contain principal-lines and wrinkles that are exclusive to that palmprint. These lines are used in the palmprint identification literature for classification purposes, but extracting these lines accurately could be very difficult and in some cases impossible. Many methods are proposed to extract lines, such as the Canny edge detection method, designing masks to compute first and second order derivatives of palm images, thresholding palm images to form binary edge images and then applying Hough transform, etc.
Here we do not aim to extract lines precisely. Instead we extract the region around each principal line and major wrinkles of the palm that contains the main characteristics of the palm. After extraction of these lines, we compute their energy in different parts of the image and use them as one set of features.
The procedure for line extraction and finding their local energy can be divided into the following steps:
\begin{enumerate}
\item	Smoothing the palm image with a Gaussian filter \vspace{-0.19cm}
\item	Extracting line edges \vspace{-0.19cm}
\item	Computing second order derivative of the palm \vspace{-0.19cm}
\item	Masking the smoothed image using the dilated version of the edge image \vspace{-0.19cm}
\item	Dividing the resulting image into T$\times$T non-overlapping blocks and computing the average energy of each block and 	      		stacking them to form the palm-line feature vector \vspace{-0.01cm}
\end{enumerate}
In the first step, the palm image must be smoothed in order to reduce the effect of noise. A common approach is to apply a Gaussian filter. The discrete 2D Gaussian filter is defined as:
\begin{equation*}
G(m,n)= \frac {1}{2 \pi\sigma^2 } \  e^{ -(\frac {(m- \mu_x)^2}{ 2\sigma_x^2}  + \frac {(n- \mu_y)^2}{ 2\sigma_y^2})}
\end{equation*}
We can convolve this filter with the palm image to derive a smoothed version of the palmprint. The result of applying the Gaussian filter to the original image, $I$, at pixel $(m,n)$ can be computed as:
\begin{equation*}
P(m,n) =  \sum_{s=-a}^{a} \sum_{t=-b}^{b} M(s,t)I(m+s,n+t)
\end{equation*}

The above filter uses parameters such as mean and variance of the filter and mask dimensions. Here we have used a zero-mean unit-variance Gaussian filter with a square mask of size $7\times7$ (a=b=3), which was chosen experimentally on a training set. Fig. 3 depicts the smoothed image.
\begin{figure}[4 h]
\begin{center}
    \includegraphics [scale=0.44] {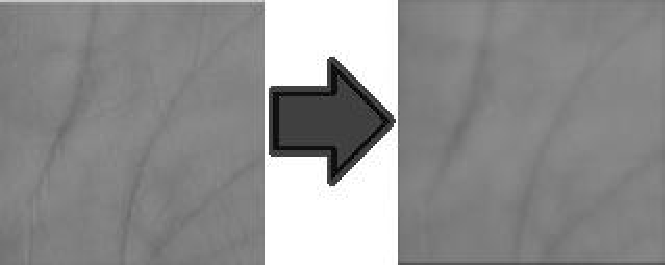}
 \end{center}
 \vspace{-0.2cm}
  \caption{Smoothed palm image }
\end{figure}

After smoothing the image, edges of the principal lines must be extracted. Here Sobel edge detection algorithm is used. 
After applying a Sobel mask, there will be some isolated points which need to be eliminated. To remove them, an isolated point removal algorithm is used. For every nonzero pixel, we look at its 8 adjacent neighbors and  if there were less than 3 nonzero neighbor pixels, we remove that pixel. The resulting edge image is shown in Fig.4.
\begin{figure}[xx2 h]
\begin{center}
    \includegraphics [scale=0.6] {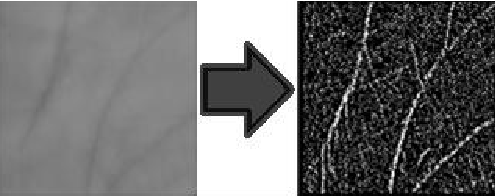}
 \end{center}
 \vspace{-0.3cm}
  \caption{Edges of the palm }
\end{figure}

In the third step, the second order derivative of the palm image is computed and all pixels except those at the local maxima are removed. The reason is that the pixels at principal line locations have darker values than their surrounding pixels, and the second order derivative often has its maximum values at these points. Fig.5 depicts the result of this step.
\begin{figure}[xx3 h]
\begin{center}
    \includegraphics [scale=0.6] {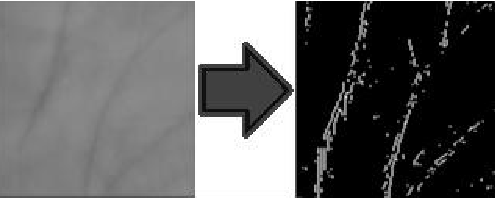}
 \end{center}
  \vspace{-0.3cm}
  \caption{Second order derivative of a palm }
\end{figure}

In the fourth step, morphological dilation is used to expand the region around edges in images of previous steps. Then the images of resulting images from steps 2 and 3 are multiplied producing an image containing major lines and wrinkles. Fig. 6 denotes the result of major line extraction. As it can be seen, not all principal lines are extracted, and some of the extracted lines are not connected. In spite of these drawbacks, this level of accuracy is good enough for our feature extraction.
\begin{figure}[xx4 h]
\begin{center}
    \includegraphics [scale=0.4] {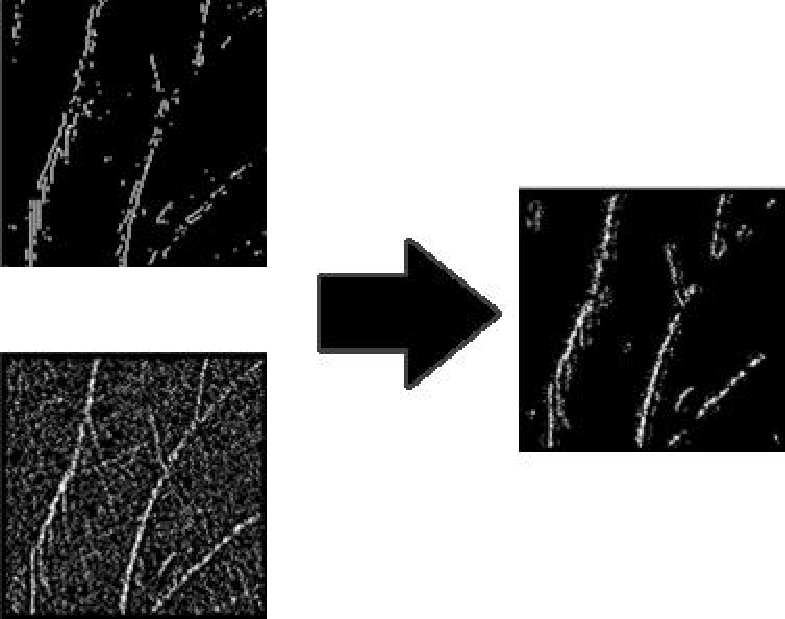}
 \end{center}
 \vspace{-0.2cm}
  \caption{Combination of edge image and second order derivative of the palm}
\end{figure}

In the fifth step, the resulting image from the last step is divided into non-overlapping blocks and average energy of each block is calculated. Then we stack the average energy of different blocks to form the palm-line feature vector of each image. 
As mentioned earlier, there are 500 different palms in the database. Let us denote the matrices representing the palmprint images by $\mathbf{ \mathit{P_1}}$ to $\mathbf{ \mathit{P_{500}}}$.
For each palm, there are 12 sample images in four spectrum bands (red, green, blue, NIR). Concatenating all of these twelve sample images in a single matrix, we get $\mathbf P_{i}$:
\begin{equation*}
 \mathbf P_{i} = [ \ \mathbf P_{i1}  \lvert  \mathbf P_{i2} \lvert \ ...\  \lvert  \mathbf P_{i,12} ]
\quad \quad \text{for}\hspace{0.5cm} i=1:500
\end{equation*}
Any of the $\mathbf P_{ij}$'s consists of four images in four different spectra. We denote the red, green, blue and NIR components with $\mathbf R_{ij}$,$\mathbf G_{ij}$,$\mathbf B_{ij}$ and $\mathbf N_{ij}$ matrices. So $\mathbf P_{ij}$ can be shown as:

\begin{equation*}
\mathbf P_{ij}= [ \mathbf R_{ij} \lvert \mathbf G_{ij} \lvert \mathbf B_{ij} \lvert \mathbf N_{ij} ]
\end{equation*}

Each of $\mathbf R_{ij},\mathbf G_{ij},\mathbf B_{ij},\mathbf N_{ij}$ is a 128 by 128 matrix, meaning $\mathbf P_{ij}$ should be a 128 by 512 matrix.
After extracting palmlines, we divide each of $\mathbf R_{ij},\mathbf G_{ij},\mathbf B_{ij},\mathbf N_{ij}$ to 4 by 4 non-overlapping blocks. So each of these matrices is divided into 1024 sub-matrices. We number these sub-matrices from 1 to 1024 as it is shown in Fig. 7.
\begin{figure}[sorting h]
\begin{center}
    \includegraphics [scale=0.6] {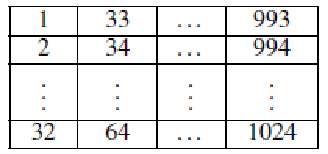}
  \caption{matrix numbering}
 \end{center}
\end{figure}

Then the feature vector of each spectrum can be derived similar to the fashion below where it is performed for the red spectrum.:
\begin{equation*}
\mathbf{f}_L(\mathbf R_{ij}) = [ f_L^1(\mathbf R_{ij}), f_l^2(\mathbf R_{ij}), ... , f_L^{1024}(\mathbf R_{ij}) ]
\end{equation*}
Where $f_L^k(\mathbf R_{ij})$ denotes the average palm-line energy in the $k$-th block (the subscript $L$ stands for line feature) and can be computed as:
\begin{equation*}
f_L^k(\mathbf R_{ij}) = \frac{1}{16 } \hspace{-0.4cm} \sum_{\stackrel{for\ all}{(m,n) \in \ \mathbf R_{ij}^{(k)}}} p(m,n)^2
\end{equation*}
and p(m,n) denotes pixel value of the image at the position $(m,n)$.

Now the feature matrices of other spectra, $\mathbf G_{ij}$,$\mathbf B_{ij}$ and $\mathbf N_{ij}$, can also be extracted using the same approach. In the end the feature vector of these four spectra are concatenated to form the final palm-line energy feature of each palmprint, which has a length of 4048 and can be shown as:
\begin{equation*}
\mathbf{f}_L(\mathbf P_{ij}) = [ \ \mathbf{f}_L(\mathbf R_{ij})  \lvert  \mathbf{f}_L(\mathbf G_{ij}) \lvert \ \mathbf{f}_L(\mathbf B_{ij}) \lvert  \mathbf{f}_L(\mathbf N_{ij}) ]
\end{equation*}

\subsection{Wavelet Features}
The discrete wavelet transform (DWT) is used in a variety of signal processing applications, such as video compression \cite{jpeg}, image denoising \cite{denoise}, image inpainting \cite{inpaint}, etc. Unlike Fourier transform, whose basis functions are sinusoids, wavelet transform is based on small waves, called wavelets, of varying frequency frequencies and limited durations. DWT can efficiently represent various signals, especially the ones that contain localized changes. 

Here, we use wavelet transform for extracting features from palmprint images.
The Daubechies-2 wavelet (db2) is applied up to 3 stages to palm images from all the four spectra, and the energy of each subband is calculated locally and used as a single feature. This method can be summarized in the following steps:
\begin{enumerate}
\item	Apply wavelet transform to each palm image up to 3 levels \vspace{-0.1cm}
\item	Divide each subband image into $T\times T$ non-overlapping blocks \vspace{-0.1cm}
\item	Compute the average energy of each block and concatenate them to form local wavelet feature vector
\end{enumerate}

Fig. 8 depicts a 3-level wavelet decomposition result.

\begin{figure}[8 h]
\begin{center}
    \includegraphics [scale=1] {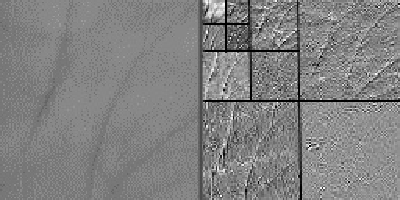}
 \end{center}
 \vspace{-0.2cm}
  \caption{Left: Palm image, Right: Its 3-level wavelet decomposition}
\end{figure}

It is noteworthy that the LLL subband is not used in the wavelet feature derivation step, but all other subbands are used.
To provide more details on how the wavelet features are derived, let us assume we want to derive the wavelet features from $j$-th sample of $i$-th person. This is shown by $P_{ij}$, which is defined as:
\begin{equation*}
\mathbf P_{ij}= [ \mathbf R_{ij} \lvert \mathbf G_{ij} \lvert \mathbf B_{ij} \lvert \mathbf N_{ij} ]
\end{equation*}
We first need to derive the wavelet transforms for all four components up to three stages: $\mathbf R_{ij},\mathbf G_{ij},\mathbf B_{ij}$ and $\mathbf N_{ij}$.
For each of them, we use 9 wavelet subbands for feature extraction (all subbands except LLL). We divide each subband into 64 blocks and calculate the average energy of any of those blocks and we put them in a $576$ dimensional vector.
This feature vector can be shown as below:
\begin{equation*}
\mathbf{f}_w({R_{ij}}) = [ \ \mathbf w_{R_{ij}}^{(1)}  \lvert  \mathbf w_{R_{ij}}^{(2)} \lvert \ ...\  \lvert  \mathbf w_{R_{ij}}^{(576)} ]
\end{equation*}
The wavelet features of $\mathbf G_{ij}$,$\mathbf B_{ij}$ and $\mathbf N_{ij}$ can also be derived in the same way as $\mathbf R_{ij}$.
Finally the feature vector of $\mathbf P_{ij}$ is formed by concatenating the features of all four spectra and is shown below:
\begin{equation*}
\mathbf{f}_w({P_{ij}}) = [ \ \mathbf{f}_w({R_{ij}})  \lvert  \mathbf{f}_w({G_{ij}}) \lvert \  \mathbf{f}_w({B_{ij}}) \  \lvert   \mathbf{f}_w({N_{ij}}) ]
\end{equation*}

\section{ Classification Algorithm for Identification}
After extracting the features of all people in the dataset, a classifier needs to be used to find the closest match of each test sample. There are different classifiers which can be utilized for this task, including neural networks \cite{NN}, support vector machines (SVM) \cite{SVM} and minimum distance classifier. In this work, minimum distance classifier is used which is popular in biometric recognition area. It basically matches a new test sample to a class with the closest distance between their feature vectors.

Suppose we want to identify the class label of a sample palmprint shown as $P_x$. We first need to find the Euclidean distance of its feature vector with that of all training samples in the dataset. 
Then the the predicted class label, $k^*$, for this palmprint can be found:
\begin{gather*}
k^*=\argmin_k \big[ dis(P_x,P_k) \big]
\end{gather*}
where $dis(P_x,P_k)$ is defined as the summation of normalized distances of line features and wavelet features:
\begin{gather*} \hspace{-0.2cm}
\mathsmaller{dis(P_x,P_k)= dis^{(n)}(\mathbf{f}_L({P_{x}}), \mathbf{f}_L({P_{k}}))+dis^{(n)}(\mathbf{f}_w({P_{x}}), \mathbf{f}_w({P_{k}}))}
\end{gather*}
where $dis^{(n)}$ denotes the normalized distance between corresponding features as shown below:
\begin{equation*}
\begin{split}
dis^{(n)}(\mathbf{f}_L({P_{x}}), \mathbf{f}_L({P_{k}})) = {\parallel  \mathbf{f}_L({P_{x}}) - \mathbf{f}_L({P_{k}})) \parallel} \ / {\bar{d}_L} \\
dis^{(n)}(\mathbf{f}_w({P_{x}}), \mathbf{f}_w({P_{k}})) = {\parallel  \mathbf{f}_w({P_{x}}) - \mathbf{f}_w({P_{k}})) \parallel} \ / {\bar{d}_w}
\end{split}
\end{equation*}
$\bar{d}_L$ and $\bar{d}_w$ are normalization factors and are determined as the following:
\begin{equation*}
\begin{split} \hspace{-0.01cm}
\bar{d}_L= \sum_{k=1}^{500} \parallel  \mathbf{f}_L({P_{x}}) - \mathbf{f}_L({P_{k}})) \parallel, \ 
\bar{d}_w= \sum_{k=1}^{500} \parallel  \mathbf{f}_w({P_{x}}) - \mathbf{f}_w({P_{k}})) \parallel
\end{split}
\end{equation*}
The reason for normalizing the distances is that the palm-line and wavelet features have different ranges and if we simply add them together, one of them might dominate the other one because of having a larger range. Despite the simplicity of the minimum distance classifier, it helps our system achieve a very high accuracy. 
We need to mention that the feature vector for the $i$-th person in the training set is defined as the average feature vector of different training images of that person and can be determined using the following equation:
\begin{equation*}
\mathbf{f}_w({P_{i}})= \frac{\sum_{j=1}^M \mathbf{f}_w({P_{ij}})}{M} 
\end{equation*}

\section{Results}
We have tested our identification algorithm on a well-known multispectral palmprint database known as PolyU database \cite{dataset}, containing 6000 palmprints captured from 500 different palms. Every palm was sampled 12 times within different times. So there are 12 $\times$ 500 = 6000 palmprint images. Every image was taken under four spectra, including red, green, blue and NIR (near infra-red). Each image has a resolution of 128 $\times$ 128. Four sample palmprints from this dataset are shown in Fig.9.
\begin{figure}[1 h]
\begin{center}
    \includegraphics [scale=0.6] {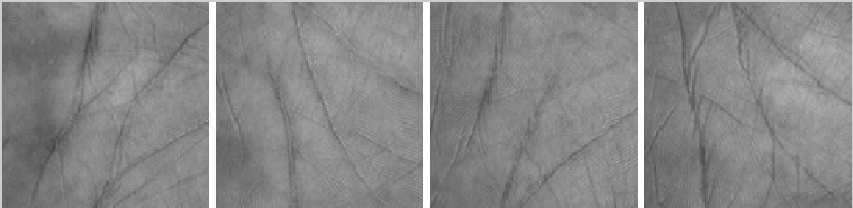}
\end{center}
\vspace{-0.2cm}
  \caption{Four sample palmprints}
\end{figure}

Correct identification takes place when a test palmprint is classified as the right person and misidentification occurs when the palmprint is classified into a category whose label is different from its actual label.
In our first experiment, half of the images from each group are used as training samples (in this experiment,  the first six samples in each palm image are selected for training) and the remaining images are used as test samples.
Among all 3000 palmprints in the database, 14 samples were misidentified, resulting in an identification accuracy of 99.53\%.

In the next experiment, to make the conditions fairer, we randomly select six palm images per person as training samples, and test on the other samples. We repeat this procedure ten times and compute the final accuracy rate by averaging on the accuracy rate of all experiments. So, the number of test samples in this experiment is 30000. In this experiment an accuracy rate of 98.8\% is obtained.

In another experiment, we examined the effect of the number of training samples on the accuracy rate (in these experiments we select the training samples randomly and repeat the experiment ten times). 
We calculated the accuracy of the proposed scheme using different numbers of training samples, ranging from 4 to 6. It was determined that reducing the number of training examples does not decrease the accuracy rate significantly. This is very desirable for the cases where the size of the training dataset is small.
The result of this experiment is shown in Table 1.

\begin{table}[h]
\centering
  \caption{Accuracy rate of the proposed method based on number of training samples}
  \centering
\begin{tabular}{|m{1.4cm}|m{1.4cm}|m{1.4cm}|m{1.4cm}|m{1.4cm}|}
\hline
Ratio of training sample & Number of test  samples & accuracy using line features & accuracy using wavelet features & accuracy using both features  \\
\hline
\ \ \ \   6/12 & \ \ \ 30000 & \ \ 94.58\% & \ \ 98.49\% & \ \ 98.88\% \\
\hline
\ \ \ \  5/12 & \ \ \ 35000 & \ \ 93.87\% & \ \ 98.25\% & \ \ 98.45\% \\
\hline
\ \ \ \  4/12 & \ \ \ 40000 & \ \ 93.22\% & \ \ 97.81\% & \ \ 98.08\% \\
\hline
\end{tabular}
\end{table}

We have also provided a comparison of the proposed scheme with two earlier approaches on this dataset, one of them being the PCA+GWR approach which uses a Gabor wavelet representation approach followed by kernel PCA, and the other one being QPCA, which uses a quaternion principal component analysis approach for multispectral palmprint recognition. This comparison is provided in Table 2.

\begin{table} [h]
\centering
  \caption{Comparison with other algorithms for palmprint recognition }
  \centering
\begin{tabular}{|m{2cm}|m{1.7 cm}|m{1.8cm}|m{1.7cm}|}
\hline
Training ratio & PCA+GWR \cite{PCA+GWR} & QPCA \cite{QPCA} & Proposed approach\\
\hline
\ \ \ \ \ \ \ 6/12 & \ \ \ 95.17\% & \ \ \ 98.13\% &  \ \ \ 98.88\%\\
\hline
\end{tabular}
\label{TblComp}
\end{table}

\section{Conclusion}
\label{SectionV}
This paper proposed a hybrid feature for palmprint recognition. The proposed hybrid feature combines two sets of features, palm-line features and wavelet features. The palm-lines are extracted by applying a Laplacian mask on the smoothed image and the wavelet features are derived from local wavelet subbands of image after applying wavelet transform up to three stages. We then use minimum distance classifier to perform template matching.
Using this hybrid feature, our algorithm is able to identify palmprints with similar line patterns and unclear palmprints.
We performed an experiment to evaluate the effect of reducing the ratio of training to test samples and it became clear that the recognition accuracy will not change too much by reducing the training ratio, which makes this algorithm desirable for cases where the amount of training data is limited. This algorithm can also be used for other biometric recognition tasks.

\section*{Acknowledgments}
The authors would like to thank biometric research group at Hong Kong Polytechnic University for providing the fingerprint dataset.

\end{document}